\documentclass[10pt,twocolumn,letterpaper]{article}

\usepackage{cvpr}
\usepackage{times}
\usepackage{epsfig}
\usepackage{graphicx}
\usepackage{amsmath}
\usepackage{amssymb}

\usepackage{subfiles}
\usepackage{multirow}
\usepackage{tabularx}
\usepackage{subfig}
\usepackage{amsfonts}
\usepackage{color}
\usepackage{booktabs}
\usepackage{pifont}

\usepackage[pagebackref=true,breaklinks=true,letterpaper=true,colorlinks,bookmarks=false]{hyperref}

\newcommand{\cblkmark}{\ding{51}}

\cvprfinalcopy 

\def\cvprPaperID{****} 

\ifcvprfinal\fi

\begin{document}

\title{Vision-Dialog Navigation by Exploring Cross-modal Memory}
%

\author{Yi Zhu$^{1}$, Fengda Zhu$^{2}$, Zhaohuan Zhan$^{3}$, Bingqian Lin$^{3}$, Jianbin Jiao$^{1}$\thanks{Corresponding Author}, 
Xiaojun Chang$^{2}$, Xiaodan Liang$^{3,4}$ \\
$^{1}$University of Chinese Academy of Sciences   
\qquad $^{2}$Monash University \\
$^{3}$Sun Yat-sen University 
\qquad $^{4}$Dark Matter AI Inc. 
}

\maketitle


\begin{abstract}

Vision-dialog navigation posed as a new holy-grail task in vision-language disciplinary targets at learning an agent endowed with the capability of constant conversation for help with natural language and navigating according to human responses. Besides the common challenges faced in visual language navigation, vision-dialog navigation also requires to handle well with the language intentions of a series of questions about the temporal context from dialogue history and co-reasoning both dialogs and visual scenes. In this paper, we propose the Cross-modal Memory Network (CMN) for remembering and understanding the rich information relevant to historical navigation actions. Our CMN consists of two memory modules, the language memory module (L-mem) and the visual memory module (V-mem). Specifically, L-mem learns latent relationships between the current language interaction and a dialog history by employing a multi-head attention mechanism. V-mem learns to associate the current visual views and the cross-modal memory about the previous navigation actions. The cross-modal memory is generated via a vision-to-language attention and a language-to-vision attention. Benefiting from the collaborative learning of the L-mem and the V-mem, our CMN is able to explore the memory about the decision making of historical navigation actions which is for the current step. Experiments on the CVDN dataset show that our CMN outperforms the previous state-of-the-art model by a significant margin on both seen and unseen environments. 
\footnote{Source code is publicly available at GitHub: \url{https://github.com/yeezhu/CMN.pytorch}}
\end{abstract}

\section{Introduction}

Powered by the recent progress in natural language processing and visual scene understanding, vision-language tasks such as Visual Question Answering (VQA) \cite{antol2015vqa,anderson2018bottom,fukui2016multimodal} and Vision-Language Navigation (VLN) \cite{fried2018speaker,anderson2018vision,mirowski2017learning,gupta2017cognitive} have been extensively explored. 
Recent works aims at developing a cognitive agent that jointly understands the natural language and visual scenes. 
However, such an agent is still far from being used in real-world applications (e.g., health care, intelligent tutoring) since it does not consider the continuous interaction with the outer environment over time. Specifically, the interaction in VQA is that the agent takes a question as input, and is required to answer a single question about a given image. The agent in VLN moves to the goal in a 3D environment following a natural language instruction.   
\begin{figure}[!t]
    \centering
    \includegraphics[width=\linewidth]{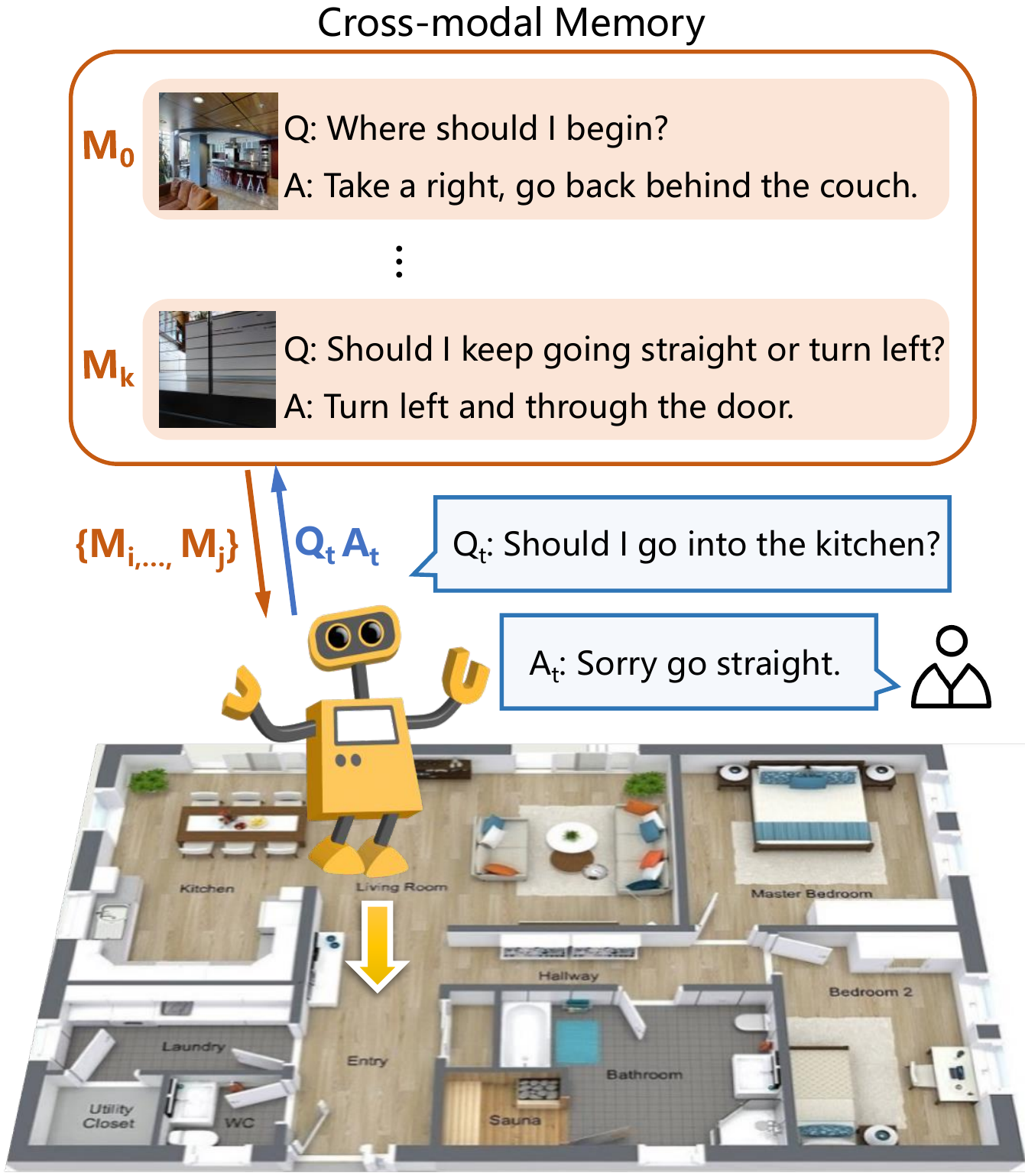}
    \caption{We propose to explore the Cross-modal Memory for vision-dialog navigation by performing co-reasoning for the memory of language interaction and visual perception. 
    }
    \label{fig:cover}
    \vspace{-0.8em}
\end{figure}
In contrast to the VQA and VLN, vision-dialog navigation \cite{thomason2019vision} is more challenging, where an agent is placed in a realistic environment and is required to find a target object by cooperating with the human using natural language dialogue. To achieve the navigation goal (i.e., find the target), the agent asks questions (e.g., Left or right from here?) to their user, saying the oracle, who knows the best actions the navigator should take. Then the navigator makes action according to the reply (e.g., Left into the bedroom) from the oracle. 
Thomason \etal simplify the cooperative vision-dialog navigation to the task of Navigation by Dialog History (NDH) where the dialogs between the navigator and the oracle are pre-annotated.
An NDH agent begins to move given an underspecified hint, which requires a series of dialogues to resolve. Previous work \cite{thomason2019vision} resolves current dialog based on dialog history without considering visual information. 
To better understand the instruction implied in the current dialog, the agent utilizes not only contextual information from the previous dialogue but also historical visually grounded information. 

In this paper, we propose a Cross-modal Memory Network (CMN) to exploit the agent memory about both the linguistic interaction with the human and the visual perception from the environment in the task of NDH. 
The CMN consists of two kinds of memory modules. 
The first module is the language memory module (L-mem), in which the dialogue histories between the navigator and the oracle are utilized to resolve the question and response at the current round. The goal of L-mem module is to give a better understanding of the instructions from the oracle who knows the best next step. 
The second module, the vision memory module (V-mem), aims at restoring the memory of visual scene during navigation. The contextualized representations generated by the L-mem are used to call back the visual memory about the places where the navigator has passed.
In CMN, the L-mem and the V-mem are used collaboratively to explore the memory about the decision making of historical navigation actions at each step, which provides a cross-modal context for the understanding of the navigation instruction indicated by the current response.

Our method has the following merits:
1) Different from the existing vision-dialog navigation method that predicts each action individually, our CMN aims to restore the memory about the previous actions.
2) CMN learns to capture the cross-modal correlations between the language and visual information and generalizes well to unseen environments.
3) CMN is simple yet effective to tackle the challenging task of NDH and significantly outperforms the previous state-of-the-art method on both seen and unseen environments on the CVDN dataset.
\section{Related Works}

\textbf{Visual Dialogue:}
The NDH agent is required to resolve the current dialog at each round based on dialog histories. Our work is related to some visual dialogue methods \cite{das2017visual, kottur2018visual, niu2019recursive, schwartz2019factor, gan2019multi,zheng2019reasoning, kang2019dual} for visual coreference resolution. 
These works learn to resolve the current sentence by exploring language attentions at word level or sentence level. ~\cite{kottur2018visual} calculates the correlation between the pronoun words and the object labels that appeared in previous dialogs. \cite{niu2019recursive, kang2019dual, gan2019multi} learn to contextualize the current question based on the attention on previous dialogs.
Different from visual dialog where the dialogs share the same visual context, the historical information about the temporal visual views is important for the NDH agent. Our method restores cross-modal memory about both the dialog history and previous visual scenes.

\textbf{Embodied Navigation:}
The problem of navigating in an embodied environment in vision and robotics has long been studied~\cite{thrun2005probabilistic, das2018embodied, gan2019look}. Despite of extensive research, embodied navigation problems remains challenging. A number of simulated 3D environments have been proposed to study navigation, such as Doom~\cite{kempka2016vizdoom}, AI2-THOR~\cite{kolve2017ai2} and House3D~\cite{wu2018building}. 
Recently, deep reinforcement learning~\cite{mnih2016asynchronous, lillicrap2016continuous, schulman2017proximal} shows its advantages in robust sequential decision making in noisy environments. Thus it is widely applied in embodied navigation. 
A number of works with deep reinforcement learning have achieve state-of-the-art results in many navigation benchmarks.~\cite{jaderberg2017reinforcement, mirowski2018learning}
However, the lack of photorealism and natural language instruction limits the application of these environments. 
Armeni \emph{et al.} propose Stanford 2D-3DS~\cite{armeni2017joint}, an embodied environment with realistic RGB-D and semantic information input. 
Anderson \emph{et al.}~\cite{anderson2018vision} propose Room-to-Room (R2R) dataset, the first Vision-Language Navigation (VLN) benchmark based on real imagery~\cite{chang2017matterport3d}. 

The VLN task has attracted widespread attention since it is both widely applicable and challenging. Earlier work~\cite{wang2018look} combined model-free~\cite{mnih2016asynchronous} and model-based~\cite{racanire2017imagination} reinforcement learning to solve VLN. Fried \emph{et al.} propose a speaker-follower framework for data augmentation and reasoning in supervised learning. In addition, a concept named ``panoramic action space" is proposed to facilitate optimization. Later work~\cite{wang2018reinforced} has found it beneficial to combine imitation learning~\cite{bojarski2016end, ho2016generative} and reinforcement learning~\cite{mnih2016asynchronous, schulman2017proximal}. The self-monitoring method~\cite{ma2019self} is proposed to estimate progress made towards the goal. Researchers have identified the existence of the domain gap between training and testing data. Unsupervised pre-exploration~\cite{wang2018reinforced} and Environmental dropout~\cite{tan2019learning} are proposed to improve the ability of generalization. 
Rich information is explored by several self-supervised auxiliary reasoning tasks \cite{zhu2019vision} to improve the visual grounding during navigation. 
The challenge of NDH task compared to VLN lies in two aspects:
1) The language instruction of VLN clearly describes the steps necessary to reach the goal while the NDH agent is given an ambiguous hint requiring exploration and dialog to resolve.   
2) The trajectory of VLN is sequential, while the NDH trajectory which consists of sub-trajectories of each dialog is hierarchical.
CMN captures the hierarchical correlation between and within sub-trajectories and explores cross-modal memory about historical actions to help better resolve the dialog.
However, current VLN methods seek to perceive the language and the visual scene sequentially.

\textbf{Vision-Dialog Navigation:}
The task of Navigation by Dialog History (NDH) was recently proposed by~\cite{thomason2019vision}, enabling the smart assistants with continuous communication and cooperation with the users via natural language and finally achieve their goal. 
An existing approach to this task follows the classical sequence-to-sequence formulation, beginning with the initial work introducing the task~\cite{thomason2019vision}. 
The actions of each step are predicted independently, this approach failed to explore the relevant information about the decision making in previous steps, and thus misled the understanding of current instruction and observation. By exploring the cross-modal memory of the agent interaction with the human and the environment, our method improves the navigation performance and make it explainable for the agent decision making procedure.

\begin{figure*}[!t]
    \centering
    \includegraphics[width=0.98\textwidth]{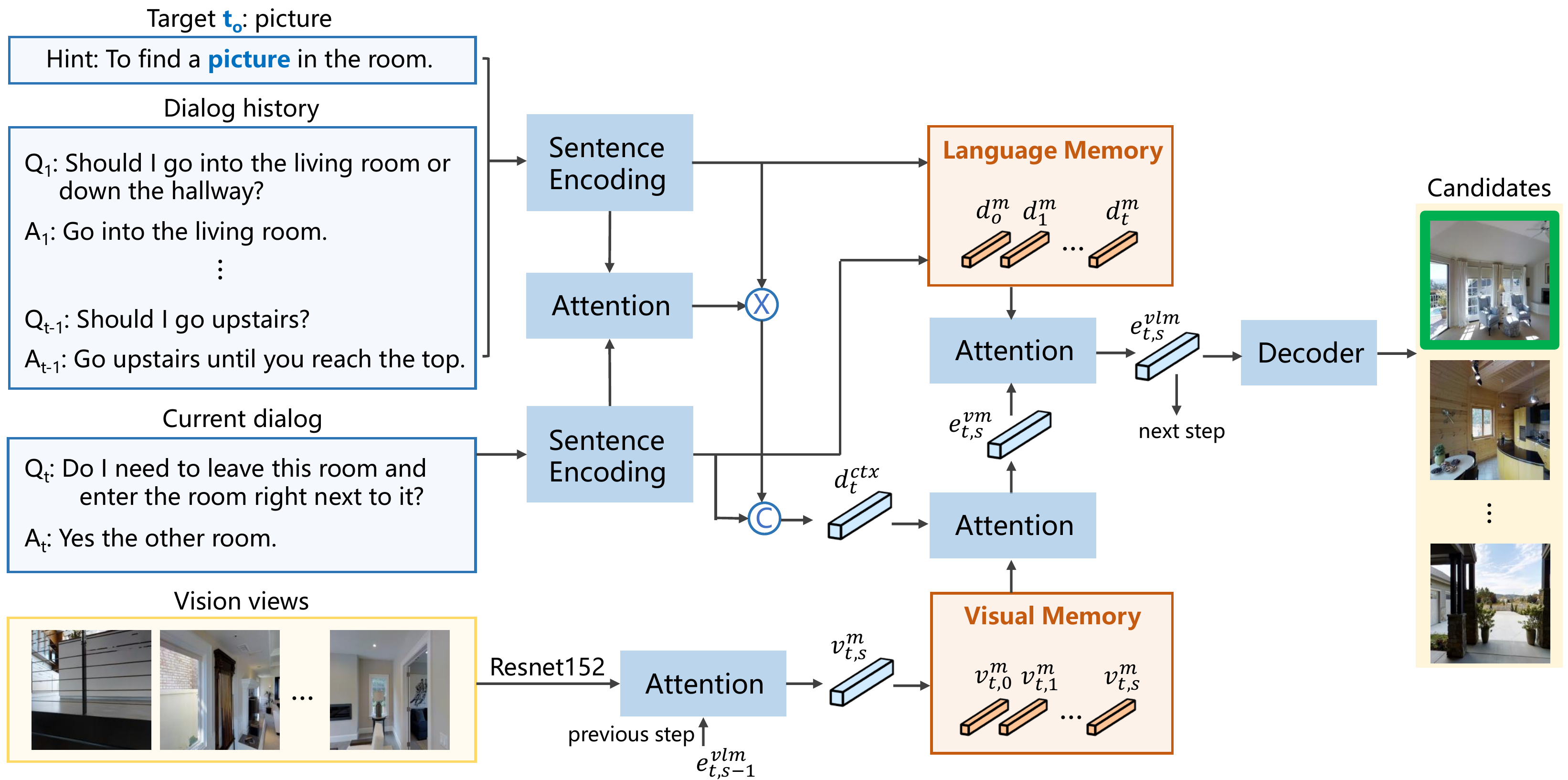}
    \caption{An overview of the Cross-modal Memory Network (CMN) for vision-dialog navigation. The panoramic views at each step are first fed to a CNN (e.g., Resnet152) to obtain panoramic representation. Then the panoramic feature of each view is fused based on the action decision of the previous step to form vision memory. The current dialog is embedded to attend on the dialog history encoding to construct contextualized representation. Following are two cross-modal attention, the language-to-vision attention takes the $d_t^{ctx}$ and visual memory as input and produce $e_{t,s}^{vm}$, then the vision-to-language attention takes the $e_{t,s}^{vm}$ and language memory as input to generate the final encoding for predicting action from the candidates.
    }
    \label{fig:framework}
\end{figure*}

\section{Method}
In this section, we briefly describe the Navigation by Dialog History (NDH) task and define the variables that will be used in the paper in Sec.~\ref{sec:problemset}. 
We introduce the feature representation for language and image in Sec.~\ref{sec:inputrep}. 
We present the Vision Memory modules (V-mem) and the  Language Memory modules (L-mem) of the proposed Cross-modal Memory Network (CMN) in Sec.~\ref{sec:vmem} and Sec.~\ref{sec:lmem}. 

\subsection{Problem Setup}
\label{sec:problemset}

According to the NDH task, the dialogs often begin with an underspecified, ambiguous instruction (e.g., Go to find the table), which requires further clarification. A dialog prompt is a tuple ($S, t_o, p_0, G_j$) contains a house scan $S$, a target object $t_o$ to be found, a starting position $p_0$, and a goal region $G_j$. At each round of communication, the navigator asks a question $Q$ and get a response $R$ from the oracle, then predict the navigation action $A$. Each sample of VDN consists of a repeating sequence $<A_0, Q_1, R_1, A_1,..., Q_k, R_k, A_k>$ for $k$ rounds of interaction.
For each dialog with prompt ($S, t_o, p_0, G_j$), a vision-dialog navigation instance is created for each of $0 \le i \le k$. The input is a hint about the target $t_o$ and a dialog history $H_t=\{D_1, ..., D_{t-1}\}$ at the $t$-th round of dialog, where $D_i = (Q_i, R_i)$.

Given the problem setup, the proposed CMN for NDH can be framed as an encoder-decoder architecture: 
(1) an encoder that explores the language memory about the historical communication $H_t$ between the navigator and the oracle, generating contextualized representation for $D_t$.  
(2) a decoder that first looks back to previous views of the navigator to help resolve the current dialog, and then converts the representation enhanced by the cross-modal memory into the navigation action space $A_t$.
Fig.~\ref{fig:framework} presents an overview of the architecture of CMN, which consists of L-mem and V-mem module. 
The L-mem learns to attend relevant previous dialogs to explore context information in a given dialog $D_t=(Q_t, A_t)$. 
The V-mem learns to recall the cross-modal memory at the previous step for the visual perception of the current scene.
\subsection{Feature Representation}
\label{sec:inputrep}

\paragraph{Language Features:} We first embed each word in the current dialog $D_t$ to {\it $\left\{w_{t,1}, ..., w_{t,T} \right\}$} by using pre-trained GloVe \cite{pennington2014glove} embeddings, where $N$ denotes the sum of the number of tokens in $Q_t$ and $R_t$. 
Then a two-layer LSTM is employed to generate a sequence of hidden states $\left\{h_{t,1}, ..., h_{t,T} \right\}$. 
The feature of each dialog $D_t$ is the the last hidden state of the LSTM $h_{t,T}$, denoted as $d_{t} \in$ $\mathbb{R}^{L}$:
\begin{equation}
\begin{split}
    \{h_{t,1},..., h_{t,N}\} =& \mathrm{LSTM(} \{w_{t,1},..., w_{t,N}\} \mathrm{)} \\
    {d_{t}} =& {h_{t,N}}
\end{split}
\label{equ:d-embed}
\end{equation} 
where $L$ is the maximal length of the dialog sentence contains a question and an answer.
Likewise, the dialog history $H_t$ is embedded following Eq.~\ref{equ:d-embed}, yielding 
{\it $\left\{d_{i} \right\}_{i=0}^{t-1}$} $\in$ $\mathbb{R}^{t \times L}$.

\paragraph{Image Features:}  
For each visual frame, we use the panoramic representation for navigation. The panoramic view is split into image patches of 36 different views, resulting in panoramic features $V_{t,s}=\{v_{t,s,i}\}, v_{t,s,i} \in \mathbb{R}^{2048}$ at the $s$-th step of round $t$, where $v_{t,s,i}$ denotes the pretrained CNN feature of the image patch at viewpoint $i$.  

\subsection{Visual Memory}
\label{sec:vmem}

We expect the navigator to make the current decision by remembering the previous cross-modal memory about the environment. Here we introduce V-mem to restore the previous cross-modal memory during navigation to help generate memory-aware representations for the current vision perception.
First, we used the final cross-modal encoding $e^{vlm}_{t,s-1}$ of the previous step $s-1$ to attend on the panoramic features $V_{t,s}$ in step $s$, the resulted memory-aware features $V^{m}_{t,s}$ depicts the correlation between previous decision and current views. 
We first project the $e^{vlm}_{t,s-1}$ and $V_{t,s}$ to $c$ dimensions and compute soft attention $A^{vis}$ as follows:
\begin{equation}
    \begin{split}
    X = f_{v}(e^{vlm}_{t,s-1}) &\odot f_{vlm}(v_{t,s,i}) \\
    A^{vis} (e^{vlm}_{t,s-1}, v_{t,s,i}) &= \sigma (X) / \sqrt{c}, \\
    \end{split}
\end{equation}
where $f_{v}(\cdot)$ and $f_{vlm}(\cdot)$ denote the two-layer multi-layer perceptrons which convert the input to $c$ dimensions. $\sigma$ represent softmax function. $\odot$ denotes hadamard product ({\it i.e.,} element-wise multiplication). 
Then we compute the memory-aware representation which contains the information about the previous action decision based on the attention $A^{vis}$ as:
\begin{equation}
    v_{t,s}^{mem} = \sum_{i=1}^s A^{vis}_{s, i} {v_{t,s,i}}.
\end{equation}
The output representations of the V-mem, $v_{t,s}^{m} \in$ $\mathbb{R}^{K}$, is calculated by applying the attention between memory about previous actions and the current views.

\subsection{Language Memory}
\label{sec:lmem}
In this section, we formally describe the Language Memory (L-mem) module. Given the current question-answer $D_t$ and the dialog history features, the L-mem module aims to attend to the memory about the most relevant dialogs in history with respect to the dialog at the current round.  
Specifically, we first compute scaled dot product attention (Attention) \cite{vaswani2017attention} in multi-head settings which are called multi-head attention. 
Let $d_{t}$ and $M_{t}$ = {\it $\left\{h_{i} \right\}_{i=0}^{t-1}$} be the current dialog and the dialog history feature vectors, respectively.  
The trainable weights $W_n^Q$, $W_n^K$ and $W^V_n \in \mathbb{R}^L \times c$ are used to project the $d_{t}$ and $M_{t}$ into features of $c$ dimension. In our experiment the dimension $c$ is set to 512. Then we calculate the attention $A^{lan}_n$ of $d_{t}$ to each element of the dialog memory $M_{t}$ as:
\begin{equation}
    \centering
    \begin{split}
        &A^{lan}_n (d_t, h_i) = \mathrm{softmax} ((d_t W_n^Q) (h_i W_n^K)^T) / \sqrt{c}, \\
        &{\hat d}_t = \mathrm {concat}_{n=1}^N \left\{ \sum_{i=0}^t A^{lan}_n(d_t, h_i) W^V_n h_i \right\}, \\
        &{\hat d}_t = \mathrm{LayerNorm}({\hat d}_t + d_t),
    \end{split}
    \label{equ:attn}
\end{equation}
where the outputs of each of the $N$ attention heads are concatenated and the contextualized representation of current dialog ${\hat d}_t$ is computed by applying a residual connection, followed by layer normalization. 
Next, the ${\hat d}_t$ is fed into a two-layer nonlinear multi-layer perceptron ($f_{lan}$), followed by a layer normalization and residual connection as:
\begin{equation}
    \begin{split}
    &{\hat d}_t = \mathrm{LayerNorm(} f_{lan}( {\hat{d}_t}) + {\hat{d}_t} ) ,\\
    &{d_t^{ctx}} = \mathrm{concat} \{ {{\hat d}_t}, {d_t}\}.
    \end{split}
    \label{equ:dctx}
\end{equation}

We then obtain the memory-aware representations by concatenating the contextual representation ${\hat d}_t$ and the original dialog representation $d_t$, denoted as $d^{ctx}_t \in$ $\mathbb{R}^{2L}$. Building on the multi-head attention mechanism, the L-mem can be stacked in multiple layers to get a high-level abstraction of the context of dialog history. 

\begin{figure}[!t]
    \centering
    \includegraphics[width=0.9\linewidth]{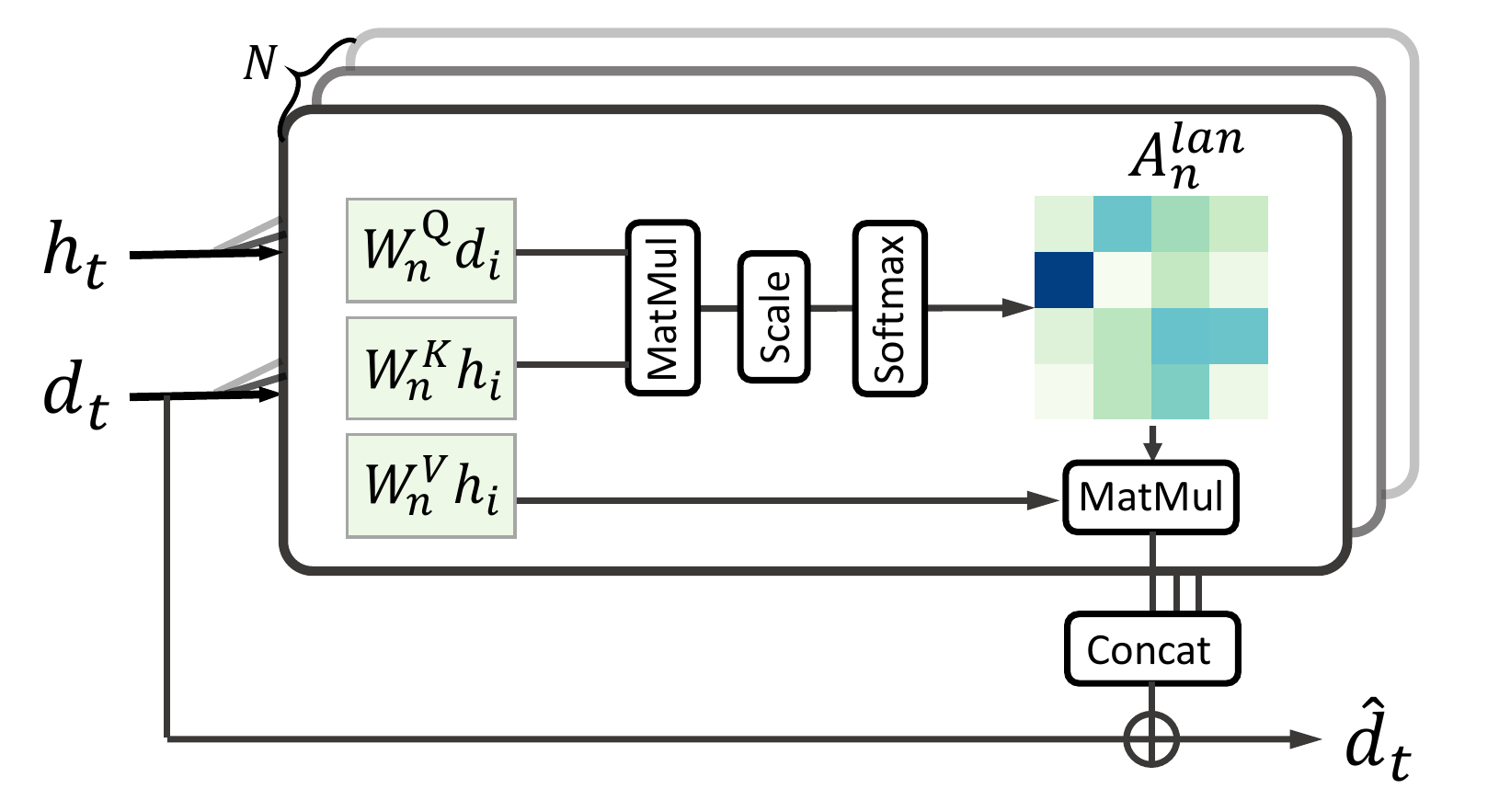}
    \caption{An illustration of the multi-head attention in Eq.~\ref{equ:attn}. $N$ represents the number of attention heads. 
    }
    \label{fig:attn}
\end{figure}

\subsection{Cross-modal Memory}
After exploring the memory of visual perception and language interactions with attention modules respectively, we further introduce cross-modal attention to explore the semantic correlation between the language and visual memory.
We first perform language-to-vision attention by leveraging the memory-aware representation of the last dialog $d^{ctx}_t$ to attend the visual memory $V^{m}_{t,s}$ via the scaled dot product attention as:
\begin{equation}
    e_{t,s}^{vm} = \mathrm{Attention}(d^{ctx}_t, \{v_{t,0}^{m},...,v_{t,s}^{m}\}).
    \label{equ:l2v}
\end{equation}
The visual memory provides supplement information about previous views, which enables better scene understanding for the navigator.    
Then we calculate vision-to-language attention to generate the final cross-modal memory encoding $e_{t,s}^{vlm}$ as:
\begin{equation}
    e_{t,s}^{vlm} = \mathrm{Attention}(e_{t,s}^{vm}, \{d_{0}^{m},...,d_{t}^{m}\}).
    \label{equ:v2l}
\end{equation}
Here the language memory is incorporated twice. The first time is in Eq.~\ref{equ:dctx} and the second time is in Eq.~\ref{equ:v2l}. The differences between the two incorporation lie in three folds. Firstly, $d^{ctx}_t$ is the concatenation of the last dialog feature $d_t$ and the attention weighted feature of previous dialog history $H_t$. So the dominate semantics derives from the last dialog $d_t$, namely, $d^{ctx}_t$ provides the contextualized information for $d_t$. In contrast, the cross-modal memory-aware representation $e_{t,s}^{vlm}$ exploit to discover the correlation between visual memory and all the existing dialogues. 
Secondly, the goal of calculating $d^{ctx}_t$ is to help the navigator better understand the current response from oracle, while the $e_{t,s}^{vlm}$ aims to learn the alignment between visual memory and language instructions, capturing the temporal correlations for better visual grounding.
Lastly, the language-to-vision attention in Eq.~\ref{equ:l2v} and the vision-to-language attention in Eq.~\ref{equ:v2l} construct a closed reasoning path between visual and language context, providing rich information of cross-modal memory for the action prediction.    

\begin{figure}[!t]
    \centering
    \includegraphics[width=1.0\linewidth]{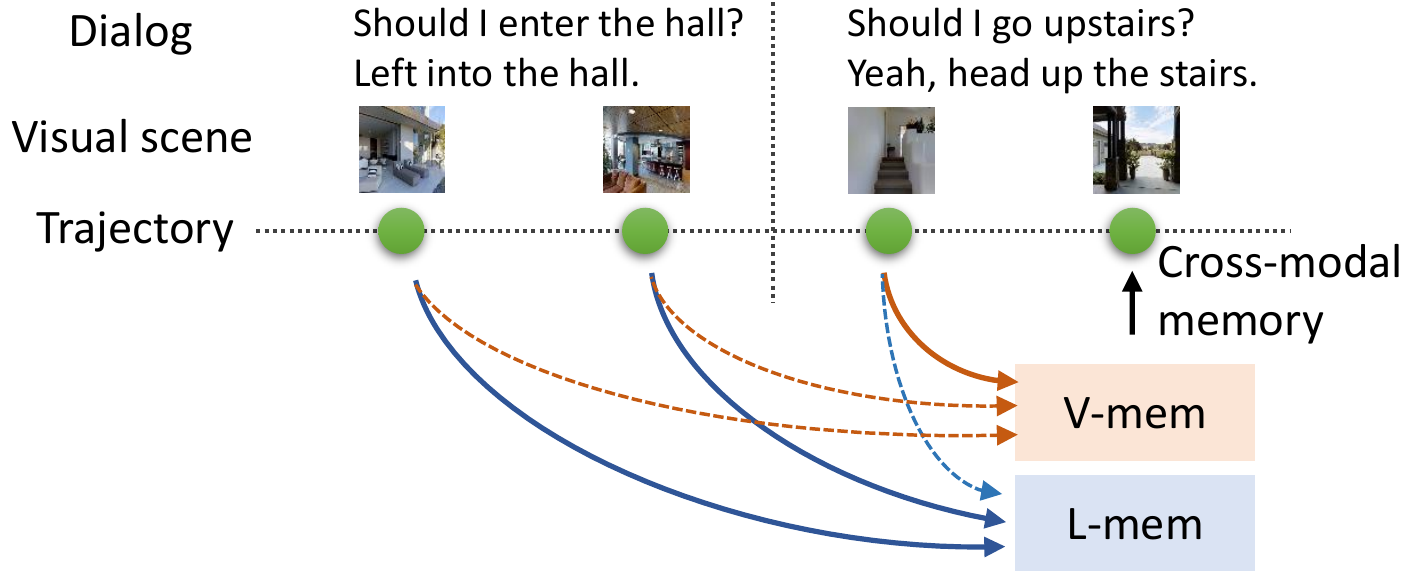}
    \caption{
    An illustration of the cooperation of the language and visual memory for each NDH instance. The language memory collects dialogs between navigator and oracle at each round, while the visual memory restores cross-modal memory of the previous navigation step. 
    }
    \label{fig:arch}
\end{figure}

In Fig.~\ref{fig:arch} we describe the cooperation of the language memory and visual memory of the navigator. The language memory is maintained for each instance of the NDH task, resolving the ambiguous instruction of oracle. In contrast, the visual memory is collected within each round to capture temporal visual cues, which could benefit visual grounding. As is shown in Fig.~\ref{fig:framework}, the action for each step is predicted based on the encoding produced by exploring the cross-modal attention between the visual and language memory.   

\subsection{Action Decoder}
By performing memory-aware reasoning with the cooperation of both language instructions and visual views, the navigator is able to better understand the historical decisions from temporal alignment between dialog history and previous views, which provide rich context information for the action prediction for the current step $s$ as:
\begin{equation}
    \begin{split}
        \hat{a}_{t,s} &= \sigma(f_{m}(e^{vlm}_{t,s})), \\
        a_{t,s} &= \mathrm{softmax}(f_a(\hat{a}_{t,s}))
    \end{split}
\end{equation}
where $f_{m}(\cdot)$ and $f_a(\cdot)$ are single-layer linear transformations to project the $e^{vlm}_{t,s}$ from $K+L$ dimension to $K$, and project the $\hat{a}_{t,s}$ from $K$ dimension to $M$ dimension which is the number of actions.
Following \cite{fried2018speaker}, we employ the panoramic action space with panoramic features of images. The agent is required to choose a candidate from the panoramic features of the visual views in the next step.

\section{Experiments}

In this section, we first introduce the experimental settings, including the CVDN dataset, evaluation metrics, and implementation details, Sec.~\ref{sec:exp-setting}. Then we compare the proposed Cross-modal Memory Networks (CMN) with previous state-of-the-art methods and several baseline models in Sec.~\ref{sec:exp-quantitative} and present ablation studies in Sec.~\ref{sec:exp-ablation}. Finally, we show the quantitative results in Sec.~\ref{sec:exp-qualitative}.

\begin{table*}[!t]
\centering
\resizebox{1.0\textwidth}{!}{
    \setlength{\tabcolsep}{0.8em}
    {\renewcommand{\arraystretch}{1.2}
        \begin{tabular}{|l|ccc|ccc|ccc|}
        \hline
\multirow{2}{*}{Method} & \multicolumn{3}{c|}{Val Seen} & \multicolumn{3}{c|}{Val Unseen} & \multicolumn{3}{c|}{Test Unseen} \\ \cline{2-10} 
& Oracle & Navigator & Mixed & Oracle & Navigator & Mixed & Oracle & Navigator & Mixed \\ \hline
Baseline (Shortest Path Agent) & 8.29 & 7.63 & 9.52 & 8.36 & 7.99 & 9.58 & 8.06 & 8.48 & 9.76 \\ 
Baseline (Random Agent) & 0.42 & 0.42 & 0.42 & 1.09 & 1.09 & 1.09 & 0.83 & 0.83 & 0.83 \\ 
Baseline (Vision Only) & 4.12 & 5.58 & 5.72 & 0.85 & 1.38 & 1.15 & 0.99 & 1.56 & 1.74 \\ 
Baseline (Dialog Only) & 1.41 & 1.43 & 1.58 & 1.68 & 1.39 & 1.64 & 1.51 & 1.20 & 1.40 \\ \hline
Sequence-to-sequence model \cite{thomason2019vision} & 4.48 & 5.67 & 5.92 & 1.23 & 1.98 & 2.10 & 1.25 & 2.11 & 2.35 \\ 
CMN (Ours) & \textbf{5.47} & \textbf{6.14} & \textbf{7.05} & \textbf{2.68} & \textbf{2.28} & \textbf{2.97} & \textbf{2.69} & \textbf{2.26} & \textbf{2.95} \\ \hline 
        
\end{tabular}
}
}
\caption{Comparison of the performance on Goal Progress (m). Different supervisions of end path are used in training. Oracle indicates planner path, Navigator indicates player path, Mixed indicates trusted path. }
\label{tab:results}
\end{table*}

\begin{table*}[!t]
\centering
\resizebox{0.92\textwidth}{!}{
    \setlength{\tabcolsep}{1.0em}
    {\renewcommand{\arraystretch}{1.2}
        \begin{tabular}{|l|cccc|cccc|}
        \hline
        \multirow{2}{*}{Method} & \multicolumn{4}{c|}{Val Seen} & \multicolumn{4}{c|}{Val Unseen} \\ \cline{2-9} 
                                & GP (m) & OSR (\%) & SR (\%) & OPSR (\%) & GP (m) & OSR (\%) & SR (\%) & OPSR (\%) \\ \hline
        Seq-to-seq \cite{thomason2019vision} & 5.92 & 63.8 & 36.9 & 72.7 & 2.10 & 25.3 & 13.7 & 33.9  \\  
        VLN Baseline \cite{fried2018speaker} & 6.15 & 58.9 & 33.0 & 69.4 & 2.30 & 35.5 & 19.7 & 45.9 \\
        CMN w/o V-mem & 6.33 & 61.3 & 30.9 & 72.3 & 2.52 & 36.7 & 20.5 & 48.4 \\  
        CMN w/o L-mem & 6.47 & 58.6 & 31.9 & 68.6 & 2.64 & 39.1 & 20.5 & 50.4 \\ \hline
        CMN (Ours) & \textbf{7.05} & \textbf{65.2} & \textbf{38.5} & \textbf{76.4} & \textbf{2.97} & \textbf{40.0} & \textbf{22.8} & \textbf{51.7} \\ \hline 
        
        \end{tabular}
}
}
\caption{Comparison of the performance on several popular benchmarks of NDH. We train a vision-language navigation method on the CVDN dataset, the performance is reported as VLN baseline. We also show the ablation studies on the L-mem and V-mem modules of our CMN.}
\label{tab:ablation}
\vspace{-0.8em}
\end{table*}

\subsection{Settings}
\label{sec:exp-setting}

\textbf{Datasets:}
We evaluate our model on the CVDN dataset which collects 2050 human-human navigation dialogs and over 7k trajectories in 83 MatterPort houses \cite{anderson2018vision}. Each trajectory corresponds to several question-answer exchanges. The dataset contains 81 unique types of household objects, each type appears in at least 5 houses and appear between 2 and 4 times per such house. Each dialog begins with an ambiguous instruction, and the subsequent question-answer interaction between the navigator and oracle will lead the navigator to find the target. 

\textbf{Evaluation Metrics:}
Following the previous works in visual language navigation and visual dialog navigation, we use four popular metrics to evaluate the proposed method from different aspects: 
(1) Success Rate (SR), the percentage of the final positions less than 3m away from the goal location. 
(2) Oracle Success Rate (OSR), the success rate if the agent can stop at the closet point to the goal along its trajectory. 
(3) Goal Progress (GP), the average agent progress towards the goal location.
(4) Oracle Path Success Rate (OPSR), the success rate if the agent can stop at the closest point to goal along the shortest path. Note that this could be different from the OSR if the shortest path is not be used for supervision (i.e., mixed path or navigator path).

\textbf{Different Supervision:}
The navigator paths in the CVDN dataset are collected from humans playing roles as the navigators, while the oracle paths are simultaneously generated by the shortest path planner. The typical supervision for the agent in the navigation task is defined by the shortest path, which is the same as the oracle path given in the CVDN dataset. However, even human demonstrations could be imperfect compared to the oracle path in realistic situations. Thus, the CVDN dataset also provides a new form of supervision called the mixed supervision path.  The mixed supervision path is defined as the navigator path when the end nodes of the navigator and the oracle are the same, and the oracle path otherwise.

\textbf{Implementation Details:}
We adopt RMSProp as the optimizer for the agent and set the learning rate to 0.0001. 
We train all agents with student-forcing for 20000 iterations of batch size 60, and evaluate validation performance every 100 iterations. 
The best performance across all epochs is reported for validation folds. The navigator moves its predicted action $\hat{a}$ at each time step. Then cross-entropy loss is applied to $\hat{a}$ and $a^*$ which is the next action along the shortest path to the target. The total training process costs 2 GPU days on a single Titan 1080Ti device. 

\subsection{Quantitative Results}
\label{sec:exp-quantitative}

\textbf{Compared Models:}
We compare our proposed CMN with several baselines and the state-of-the-art method:
(1) The Shortest Path Agent takes the shortest path to the supervision goal at inference time and represents the upper bound navigation performance for an agent.
(2) The Random Agent chooses a random heading and walks up to 5 steps forward (as in \cite{anderson2018vision}). 
(3) The Vision Only baseline where the agent considers visual input with empty language inputs.
(4) The Dialog Only baseline where the agent considers language input with zeroed visual features.
(5) The sequence to sequence model proposed in \cite{thomason2019vision} where the historical dialogs are concatenated to form a single instruction as in visual language navigation models \cite{anderson2018vision}.

\textbf{Comparison to previous methods:}
As is shown in Tab.~\ref{tab:results}, our proposed CMN outperforms the previous state-of-the-art method \cite{thomason2019vision} on the Goal Progress (m) with different supervisions ( e.g., planner path (Oracle), player path (Navigator) and trusted path (Mixed)), demonstrating the ability of our method to grounding visual elements in the explored environments. When evaluated on the Val Unseen and Test Unseen data, the gap between our CMN and the seq-to-seq method is also significant, showing that CMN generalizes well for unexplored environments. 

\subsection{Ablation Study}
\label{sec:exp-ablation}
We ablate the V-mem and L-mem module on the Val sets in Tab.~\ref{tab:ablation} and the Test set in Tab~\ref{tab:test-ablation}.  
\begin{table}[!t]
\centering
\resizebox{0.8\linewidth}{!}{
    \setlength{\tabcolsep}{0.8em}
    {\renewcommand{\arraystretch}{1.0}
        \begin{tabular}{|ccc|c|}
        \hline
        V-mem & L-mem & VL-mem & Goal Process (m) \\ \hline
         \cblkmark & \cblkmark & \cblkmark & 2.95 \\
         & \cblkmark & \cblkmark & 2.74 \\ 
         \cblkmark &  & \cblkmark & 2.04 \\ 
         \cblkmark & \cblkmark &  & 2.54 \\ \hline 
        \end{tabular}
}
}
\caption{Ablation studies on different types of memory information for our proposed CMN, including visual memory, language memory, and cross-modal memory.}
\label{tab:test-ablation}
\end{table}

\textbf{Baselines:}
In the first row, we conduct a baseline from VLN by directly using the concatenated dialog history as the language input. The difference between the VLN baseline and the Sequence-to-sequence model are two folds. First, the Seq-to-seq model uses a $1 \times 2048$ feature vector to represent each panoramic image, while the dimension of the visual feature used in baseline VLN is $ 1 \times 36 \times 2048$ for all 36 views of the panoramic image. Second, the action space in the Seq-to-seq model is the low-level visuomotor space, where the predictions of actions are $3$-d logits.
In contrast, the baseline VLN uses panoramic action space, where the agent has a complete perception of the scene and directly performs high-level actions. Our framework is built based on the baseline VLN with both the panoramic vision features and the panoramic action space.  

\textbf{Effect of different memory module:}
We disable the V-mem module by directly averaging the visual features of each panoramic view. 
In Tab.~\ref{tab:ablation} and Tab.~\ref{tab:test-ablation} we can see that the performance dropped when the model lost visual memory about previous navigation steps. 
The reason why we use averaged feature here is that the averaged features are more confused and useless than the last memory features since closer memory is more correlated to the current state than earlier memory, which helps us disable most of the functionality of memory modules in ablation studies.

To remove the L-mem module, we replace the memory-aware representation of the language interactions by the word-level context within the last question-answer sentences. It can be seen in Tab.~\ref{tab:ablation} and Tab.~\ref{tab:test-ablation} that the performance of our method dramatically decreases when discarding the language memory module (L-mem), indicating that the language memory is crucial for the understanding of the oracle instructions.
In Tab.~\ref{tab:test-ablation}, we also set the output of our encoder ($e_{t,s-1}^{vlm}$ Fig.~\ref{fig:framework}) to zero values to eliminate the cross-modal memory context (VL-mem) from the previous navigation step. The VL-mem can be regarded as a high-level abstraction of historical navigation and represents rich information about the previous action decision made by the agent. The results indicate that the performance of our model would drop when the cross-modal memory is lost.

\textbf{Discussion about the temporal order in memory: }
As is shown in Fig.~\ref{fig:framework}, CMN predicts the action at step $s$ by restoring the cross-modal memory $e^{vlm}_{t,s-1}$, which represents the memory about the decision making of the previous navigation action at step $s-1$. 
From this point of view, the information about the temporal order of the cross-modal memory is implicitly embedded in our CMN. 
We further consider a more explicit way that directly concatenates the embedding of the order and the memory features. The performance is comparable to the implicit way.

\subsection{Qualitative Results}
\label{sec:exp-qualitative}
\begin{figure*}[!t]
    \centering
    \includegraphics[width=0.9\textwidth]{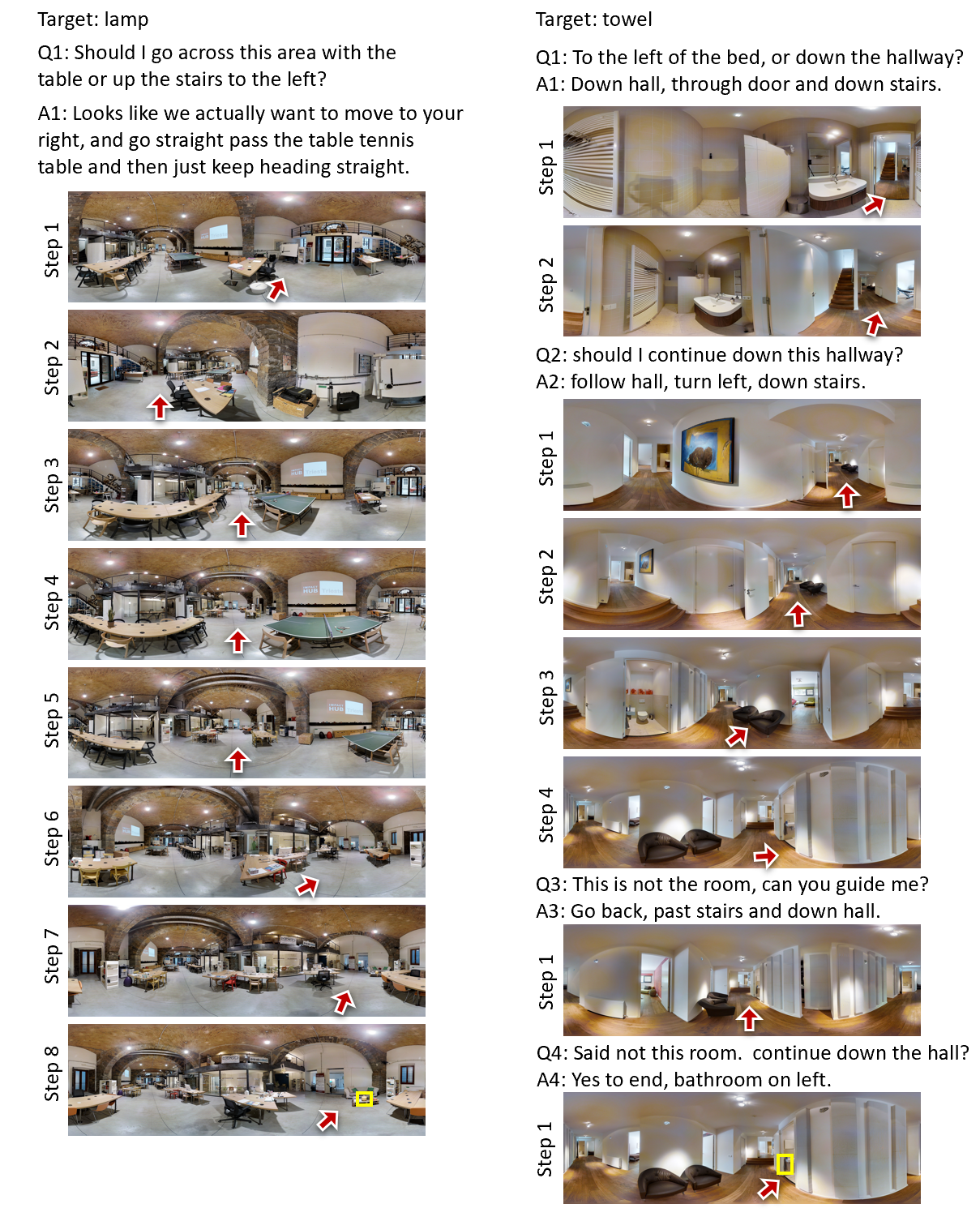}
    \caption{Examples of vision dialogue navigation using our proposed Cross-modal Memory Network. The red arrows indicate the predicted actions and the yellow boxes indicate the targets. Best viewed in color.
    }
    \label{fig:vis}
    \vspace{-0.8em}
\end{figure*}

To see how our proposed CMN performs the visual dialog navigation task, we visualize two qualitative examples in Fig.~\ref{fig:vis}. The first example contains one round of dialogue with eight steps of navigation. We can see that the agent successfully reaches the target with a comprehensive understanding of the natural language response from oracle, indicating that our method is also compatible with the VLN task.
In the second example, there are five rounds of dialogue between the navigator and the oracle. In the first round of communication, the navigator moves two steps down to the hallway, which requires the navigator to correctly recognize visual elements, including bed, hallway, and stairs while understanding language instructions. In the third round of communication, the oracle suggests the navigator to ``go back''. Our CMN explores to better understand this instruction by referring it to dialog history to resolve the specific meaning, that is, the inverse navigation operation of the previous steps. In step 1 of the third round, the navigator returns back to the hall. Finally, it finds the target ``towel''. 
Since our proposed Cross-modal Memory could help restore the visual and language memory of previous steps and interactions, the navigator can resolve the ambiguous instruction ``go back'' and thus return to the previous location. 
 
\section{Conclusion}
In this work, we propose the Cross-modal Memory Network (CMN) to tackle the challenging task of visual dialogue navigation by exploring cross-modal memory of the agent. The language memory can help the agent better understand the responses from the oracle based on the communication context. The visual memory aims to explore visually grounded information on the previous navigation path, providing temporal correlations for the views. Benefiting from the collaboration of both visual and language memory, CMN is proved to achieve constant improvement over popular benchmarks on visual dialogue navigation, especially when generalizing to the unseen environments. 

\textbf{Acknowledgement.} 
This work was supported in part by the National Natural Science Foundation of China (NSFC) under Grant
No.U19A2073, No.61976233, No.61836012, and No.61771447.


{\small
\bibliographystyle{ieee_fullname}
\bibliography{egbib}
}

\end{document}